\documentclass[journal]{IEEEtran}
\usepackage{cite}
\usepackage{amsmath,amssymb,amsfonts}
\usepackage{algorithmic}
\usepackage{graphicx}
\usepackage{textcomp}

\usepackage{amsmath,amssymb} 

\usepackage{caption}

\usepackage[american]{babel}
\usepackage{microtype}

\usepackage{verbatim}
\usepackage{bm}
\usepackage{graphicx}
\usepackage{subfigure}

\usepackage{makecell}
\usepackage{multirow}
\usepackage{url}

\usepackage{multirow}

\usepackage{makecell}

\usepackage{color}
\usepackage{booktabs} 
\usepackage{colortbl} 
\definecolor{revisionred}{rgb}{1,0,0} 

\def\BibTeX{{\rm B\kern-.05em{\sc i\kern-.025em b}\kern-.08em
    T\kern-.1667em\lower.7ex\hbox{E}\kern-.125emX}}

\begin{document}

\title{Unsupervised Domain Adaptation Network with Category-Centric Prototype Aligner for\\ Biomedical Image Segmentation}

\author{Ping~Gong*,
        Wenwen~Yu*,
        Qiuwen~Sun,
        Ruohan~Zhao,
        and Junfeng~Hu
    \thanks{The * indicates equal contribution}
    \thanks{P. Gong, W. Yu, Q. Sun, and J. Hu are with the School of Medical Imaging, Xuzhou Medical University, Xuzhou, 221000, China.}
    \thanks{R. Zhao is with the Department of Computing, The Hong Kong Polytechnic University, Kowloon 999077, Hong Kong.}
    \thanks{Email at: W. Yu \textit{yuwenwen62@gmail.com}}
}

\maketitle

\begin{abstract}
With the widespread success of deep learning in biomedical image segmentation, domain shift becomes a critical and challenging problem, as the gap between two domains can severely affect model performance when deployed to unseen data with heterogeneous features. To alleviate this problem, we present a novel unsupervised domain adaptation network, for generalizing models learned from the labeled source domain to the unlabeled target domain for cross-modality biomedical image segmentation. Specifically, our approach consists of two key modules, a conditional domain discriminator~(CDD) and a category-centric prototype aligner~(CCPA). The CDD, extended from conditional domain adversarial networks in classifier tasks, is effective and robust in handling complex cross-modality biomedical images. The CCPA, improved from the graph-induced prototype alignment mechanism in cross-domain object detection,  can exploit precise instance-level features through an elaborate prototype representation. In addition, it can address the negative effect of class imbalance via entropy-based loss.  Extensive experiments on a public benchmark for the cardiac substructure segmentation task demonstrate that our method significantly improves performance on the target domain.
\end{abstract}


\section{Introduction}
\label{sec:introduction}
Deep neural networks have achieved remarkable success in recent years; a variety of challenging medical imaging problems, such as biomedical image segmentation~\cite{Taghanaki2020DeepSS,Shelhamer2017FullyCN}, have witnessed breakthroughs, when a large  quantity of labeled data are used, and the training and testing data are sampled from the sample distribution\cite{iek20163DUL}. However, deep neural networks deployed in real-life applications usually suffer from the domain shift problem~\cite{BenDavid2009ATO}. In biomedical imaging scenarios, this problem is even more obvious, as biomedical images have very different characteristics when they are acquired with different acquisition parameters or modalities~\cite{Dou2018UnsupervisedCD,Dou2019PnPAdaNetPA}, such as magnetic resonance imaging (MRI) and computed tomography (CT). In addition, manual annotation in the field of biomedical images is a time-consuming and expensive task. Compared to natural images, domain adaptation is more challenging on cross-modality biomedical images.

To address these problems, unsupervised domain adaptation~(UDA) has been intensively studied to generalize the well-trained models on unlabeled target data, which seeks to transfer knowledge from labeled training data ~(source domain) to test data ~(target domain). To obtain domain-invariant feature representations across both domains, common methods for domain adaptation can be roughly divided into two types: 1)~measures for minimizing an explicitly defined distance~\cite{Yan2017MindTC,Long2015LearningTF}, and 2)~applying adversarial training to align the latent feature space~\cite{Ganin2015UnsupervisedDA, Pei2018MultiAdversarialDA}. 

\begin{figure*}[htb]
\centering
\includegraphics[width=0.99\textwidth]{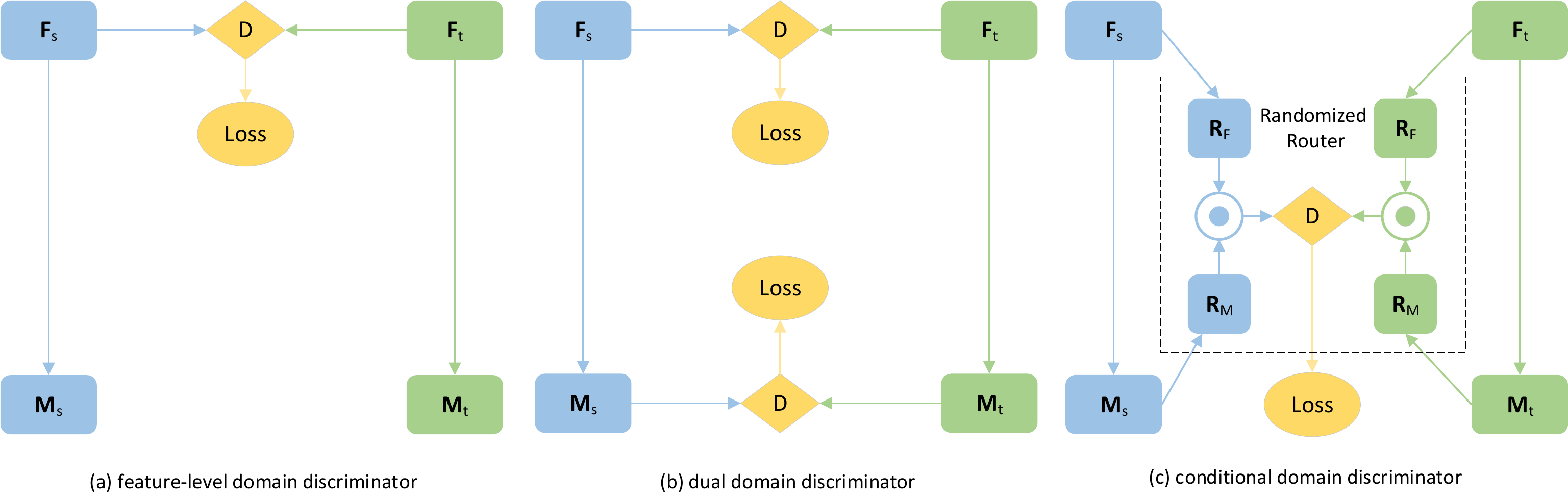}
\caption{Frameworks of different domain discriminator.  (a)~feature-level domain discriminator based method.  (b)~dual-domain discriminator (feature-level and instance-level)-based method. (c)~our proposed models named the conditional domain discriminator.}
\label{fig:discriminator}
\end{figure*}

For instance, Gretton \textit{et. al}.~\cite{Gretton2012AKT} minimized the maximum mean discrepancy (MMD) distance between the source and target domains. Long \textit{et al}.~\cite{Long2015LearningTF} proposed the multikernel MMD distance. Other works based on adversarial training employed a domain classifier to facilitate domain invariance on the input level~\cite{Sankaranarayanan2018GenerateTA}, feature level~\cite{Ganin2015UnsupervisedDA}~(Figure~\ref{fig:discriminator}(a)), output level and their combinations~\cite{Dou2019PnPAdaNetPA}~(Figure~\ref{fig:discriminator}(b)). Nevertheless, how to balance the ratio between multilevel discriminators requires empirical design and time for UDA and can severely affect the domain adaptation performance. Furthermore, multilevel discriminators may come at a contradiction in the adversarial adaptation procedure given the complex combinations of cross-modality between domains. In addition, domain classifiers tend to capture global-level discrepancy features across both domains, ignoring distinct modal information of instances in the cross-modality scenario, especially when neighboring structures contain unclear boundaries or relatively homogeneous tissues\cite{Dou2019PnPAdaNetPA}, and combining the feature-level and instance-level alignment of the source and target domains makes it difficult to perfectly address domain shift~\cite{Xu2020CrossDomainDV, Pan2019TransferrablePN, Saito2019SemiSupervisedDA, Xie2018LearningSR}.

\begin{figure*}[htb]
\centering
\includegraphics[width=0.99\textwidth]{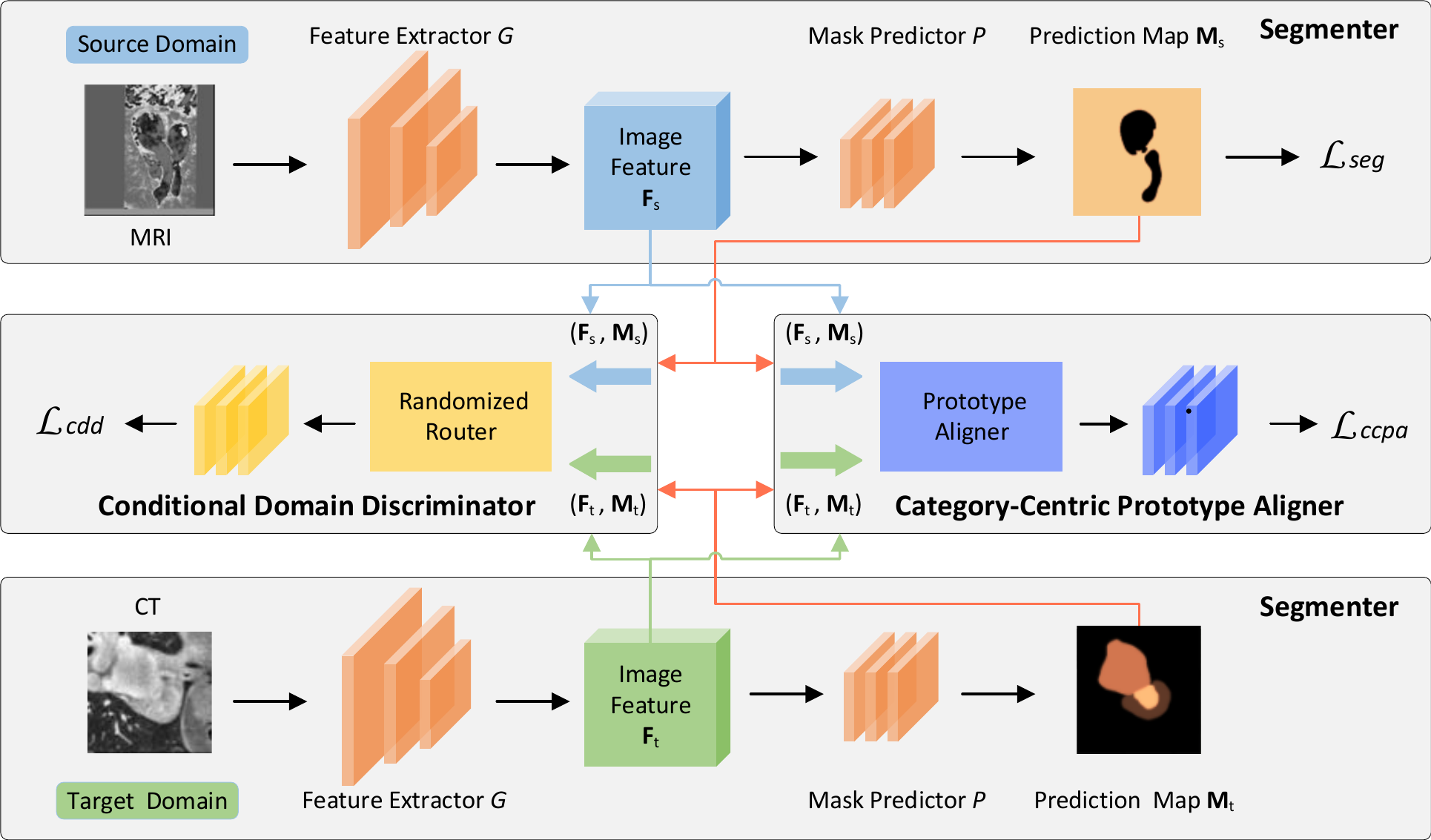}
\caption{Overview of the framework. The segmenter is shared for both the source and target domains. The conditional domain discriminator and category-centric prototype aligner are trained from both source and unlabeled target data. During inference, the target domain is predicted by the segmenter.}
\label{fig:overall}
\end{figure*}

Motivated by these problems, we propose the unsupervised domain adaptation network with  category-centric prototype aligner for biomedical image segmentation illustrated in Figure~\ref{fig:overall}. Specifically, we introduce two key components, a conditional domain discriminator~(CDD) and a category-centric prototype aligner~(CCPA). Our method incorporates the CDD module inspired by~\cite{Long2018ConditionalAD} into the existing architecture to model the multimodal information and joint distributions of multilevel features via a randomized router, which makes it easier to fully capture multiplicative interactions between feature representation and segmentation prediction for domain adaptation. In CCPA, we extend category-level prototype alignment to cross-domain segmentation tasks, inspired by~\cite{Xu2020CrossDomainDV, Pan2019TransferrablePN}, to guarantee the discriminability by exploiting more precise instance-level features. In addition, it can mitigate the negative effect of class imbalance on domain adaptation via an entropy-based loss to control the process of adaptation during the training phase.

The main contributions of this paper can be summarized as follows:
\begin{itemize}
\item{In this paper, we present a novel framework for UDA, which is more effective and robust in handling complex cross-modality biomedical image segmentation. It efficiently balances multi-level features discriminator without contradiction that is crucial for UDA.} 

\item{We introduce CCPA module into the image segmentation framework, which exploits more precise instance-level features. In addition, it also can mitigate the negative effect of class-imbalance via entropy-based loss.}

\item{We conduct experiments on one public benchmark for cardiac substructure segmentation tasks and show that our method significantly improves performance on the target domain.}
\end{itemize}

\section{Related Work}
The purpose of UDA is to generalize the model learned from the labeled source domain to the unlabeled target domain. A large number of adaptive methods have been proposed from different perspectives, including input-level, feature-level, output-level adaptation, and their combinations. In academia, UDA can be divided into two categories: 1)~measures for minimizing a specific domain discrepancy metric, and 2)~applying adversarial training to align the latent features space. In this section, we will give a brief review of related works in both areas. A more detailed review of UDA for image segmentation can be found in~\cite{Toldo2020UnsupervisedDA, Sun2015ASO, Wang2018DeepVD, Patel2015VisualDA,Chen2020DeepLF}. 

In the field of UDA, early research mainly focused on aligning the distributions of feature space by minimizing distance measurements. For example, Tzeng \textit{et al}.~\cite{tzeng2014deep} used the MMD distance as a minimized target between the source and target domains. Shen \textit{et al}.~\cite{shen2017wasserstein} learned domain-invariant feature representations via Wasserstein distance guided representation learning. Long \textit{et al}.~\cite{Long2015LearningTF} presented the multikernel MMD distance based on MMD to reduce the domain discrepancy. Yan \textit{et al}.~\cite{Yan2017MindTC} further extended the work and employed weighted MMD with a task-specific loss for domain adaptation. Ding \textit{et al}.~\cite{Ding2020AdaptiveEF} proposed an adaptive exploration method to maximize the distances of all target images, minimizes distances of similar target images, and address the domain-shift problem for person re-identification~(re-ID) in an unsupervised manner. Fan \textit{et al}.~\cite{fan2018unsupervised} proposed a progressive unsupervised learning method to transfer pretrained deep representations to unseen domains based on clustering to improve the re-ID accuracy and produced CNN models with high discriminative ability. Nevertheless, both~\cite{Ding2020AdaptiveEF,fan2018unsupervised} are aimed at the topic of re-ID. In addition, the difference between~\cite{Ding2020AdaptiveEF,fan2018unsupervised} and our method is that~\cite{Ding2020AdaptiveEF} minimizes distances between one image and its neighbors and maximizes distances between one image and other images at the feature level, and~\cite{fan2018unsupervised} performs clustering at the feature level. However, our method is not only performs conditional domain discriminator at the  feature level and instance level, but also aligns category-centric prototype at the instance-level.

More recently, with the advent of generative adversarial networks~(GAN)~\cite{Goodfellow2014GenerativeAN}, another line of research is based on adversarial training. For instance, Ganin \textit{et al}.~\cite{Ganin2015UnsupervisedDA} aligned the distributions of features across the two domains accomplished through standard backpropagation training via a simple new gradient reversal layer. Tzeng \textit{et al}.~\cite{Tzeng2017AdversarialDD} introduced a more flexible adversarial learning framework with united weight sharing to address the problem of domain shift. With the widespread success of CycleGAN~\cite{Zhu2017UnpairedIT} in unpaired image-to-image transformations, many previous image adaptation efforts were based on modified CycleGAN with applications in both natural datasets~\cite{Hoffman2018CyCADACA, Russo2018FromST} and medical image segmentation~\cite{Zhang2020UnsupervisedXI,Chen2019SynergisticIA, Chen2020UnsupervisedBC}. Overall, our framework belongs to the latter category based on adversarial learning.

For biomedical image segmentation applications, as the common cross-modality, interscanner, and different imaging protocols vary, domain shift has become a critical and challenging problem in biomedical image segmentation. Ghafoorian \textit{et al}.~\cite{Ghafoorian2017TransferLF} reduced the required number of labels in the target domain via transfer learning for brain MRI lesion segmentation tasks. Opbroek \textit{et al}.~\cite{Opbroek2015TransferLI} proposed four transfer classifiers to deal with differences in distributions between training and target data and improved performance over supervised learning for segmentation across scanners and scan protocols. Nevertheless, these methods require extra annotations of target data. Instead, unsupervised domain adaptation is more desirable due to the requirements of zero additional target domain labels. Recent literature shows the great potential of unsupervised domain adaptation for biomedical image segmentation. For example, Kamnitsas \textit{et al}.~\cite{kamnitsas2017unsupervised} made the earliest attempts to align feature distributions with an adversarial loss for unsupervised domain adaptation cross-protocol MRI segmentation and achieved promising adaptation performance. Degel \textit{et al}.~\cite{degel2018domain} and Zhang \textit{et al}.~\cite{Zhang2018MultiInputAD} combined regularization with adversarial training and obtain better adaptation results on ultrasound datasets and cardiac MRI segmentation, respectively. Wang \textit{et al}.~\cite{Wang2019PatchBasedOS} presented a novel patch-based output space adversarial learning framework to jointly and robustly segment the optic disc and optic cup from different fundus image datasets and achieved effective feature alignment. However, these works did not aim at the topic of unsupervised domain adaptation for cross-modality biomedical images. In the field of cross-modality biomedical segmentation, Dou \textit{et al}.~\cite{Dou2018UnsupervisedCD,Dou2019PnPAdaNetPA}  employed adversarial learning to adapt the early-layer feature distributions while the higher-layer features are fixed. However, their method needs comprehensive empirical studies and design to determine the optimal adaptation depth. Chen \textit{et al}.~\cite{Chen2019SynergisticIA, Chen2020UnsupervisedBC} used deeply synergistic image and feature alignment for unsupervised bidirectional cross-modality adaptation segmentation and achieved better domain performance, but it has a complex training process due to the essence of GAN is used.

The most related works to our method are~\cite{Xu2020CrossDomainDV,Long2018ConditionalAD} based on the graph-inducted prototype alignment mechanism and conditional domain adversarial networks, but still differ from our method in several aspects. These works are not used for the field of unsupervised domain adaptation for cross-modal biomedical image segmentation. More specifically, the difference between~\cite{Long2018ConditionalAD} and our proposed CDD module is that conditional adversarial domain adaptation is a common operation in the classifier tasks but rarely in segmentation tasks, especially in the field of unsupervised domain adaptation on cross-modal medical image segmentation. The difference between~\cite{Xu2020CrossDomainDV} and the proposed CCPA is that~\cite{Xu2020CrossDomainDV} is only suitable for object detection tasks because it performs graph-induced prototype alignment on the bounding-box level. However, the proposed CCPA module extends~\cite{Xu2020CrossDomainDV} to the pixel-level for segmentation tasks. We use a learnable graph attention layer to obtain the adjacency matrix of the graph at the pixel-level instead of calculating weights between region proposals at the bounding-box level.

\section{Method}
In this section, we provide a detailed description of our proposed method for UDA in biomedical image segmentation. Figure~\ref{fig:overall} gives the overall architecture of our method, which contains 3 modules: 

\begin{itemize}
\item{\emph{Segmenter}}: This module consists of the feature extractor $G$, and the mask predictor $P$. The feature extractor $G$ encodes source/target domain~(MRI/CT) images using CNN to obtain image features $\mathbf{F}_s$/$\mathbf{F}_t $. $\mathbf{F}_s$ and $\mathbf{F}_t$ represent semantic information of the source domain and target domain in the latent feature space, respectively. Then the mask predictor $P$ uses image features $\mathbf{F}_s$/$\mathbf{F}_t$  as input to predict the segmentation map $\mathbf{M}_s$/$\mathbf{M}_t$ of the source domain and target domain individually.

\item{\emph{Conditional Domain Discriminator}}: This module can capture the latent relation between image feature $\mathbf{F}$ and prediction map $\mathbf{M}$  through a randomized router, in which image features interact with the prediction map. In addition, it can capture the cross-covariance between image feature $\mathbf{F}$  and prediction map $\mathbf{M}$ to improve the transferability.

\item{\emph{Category-Centric Prototype Aligner}}: This module performs category-level prototype alignment to guarantee discriminability by exploiting more precise instance-level features. In addition, it can mitigate the negative effect of class imbalance on domain adaptation via an entropy-based loss to control the process of adaptation during the training phase. Thus, CCPA can obtain the capacity of transfer from the source domain to the target domain through UDA. 
\end{itemize}

To ease understanding, our full model is described in parts. First, we begin by introducing the notation used in this paper in Section~\ref{sec:notation}.  Our segmenter representation is described in Section~\ref{sec:segmenter}; then, the proposed CDD module mechanism is described in Section~\ref{sec:cdd}. Finally, Section~\ref{sec:ccpa} shows how the CCPA module works.

\subsection{Notation}
\label{sec:notation}

In UDA, given $N^s$ labeled examples of the source domain and  $N^t$ unlabeled examples of the target domain, they are denoted by $\mathcal{D}_s =\left\lbrace( x_1^s, y_1^s),\ldots, (x_{N^s}^s, y_{N^s}^{s}) \right\rbrace$ and $\mathcal{D}_t=\left\lbrace x_1^t,\ldots, x_{N^t}^t \right\rbrace$, respectively, where $x_i^s $ is the image for the $i$-th sample of the source domain and follows distributions $\mathbb{P}_s$, and $y_i^s $ represents its corresponding segmentation one-hot label. Similarly, $x_i^t$ follows distributions $\mathbb{P}_t$, and the i.i.d. assumption is violated as $\mathbb{P}_s \ne \mathbb{P}_t$.

For ease of notation, we directly use $x^s$ and $y^s$ to represent the sample and label from the source domain, where we omit the subscript index $i$ in the following subsections.

\subsection{Segmenter}
\label{sec:segmenter}
As shown in Figure~\ref{fig:overall}, the top position in the diagram is the segmenter module, which contains two steps for the source domain. Specifically, image sample $x^s$ is first forwarded into the feature extractor $G$, yielding image features $\mathbf{F}_s \in \mathbb{R}^{ H\times W \times d }$ in latent feature space, where $ H, W, d $ denotes the height and width of the output map, and the dimension of feature embedding, respectively. Next, to perform segmentation on image features, the mask predictor $P$ learns to create a mapping from image features to the label space $\mathcal{Y}^s$ by conducting supervised learning. Formally, prediction map $\mathbf{M}_s \in \mathbb{R}^{ H\times W \times N_{c} }$, where $ N_c $ denotes the number of classes, is defined as follows:
\begin{equation}
\begin{cases}
\mathbf{M}_s = P\left(\mathbf{F}_s~;  \Theta_\text{p}\right) \,,\\
\mathbf{F}_s = G\left(x^s~;  \Theta_\text{g}\right) \,,
\end{cases}
\end{equation}
where $\Theta_\text{p}$ and  $\Theta_\text{g}$ represent the learned parameters of the mask predictor and the feature extractor, respectively. Similarly,  as shown in the bottom position of Figure~\ref{fig:overall}, the target domain sample $x^t$ also has the same procedure to generate $\mathbf{F}_t \in \mathbb{R}^{ H\times W \times d }$ and $\mathbf{M}_t \in \mathbb{R}^{ H\times W \times N_{c} }$ during the domain adaptation phase, which is also fed into CDD and CCPA. In addition,  $\Theta_\text{p}$ and  $\Theta_\text{g}$ are shared parameters for both the source and target domains. In practice, we apply dilated residual blocks~\cite{Yu2017DilatedRN} to the feature extractor $G$ to obtain a large receptive field and preserve the spatial acuity of feature maps. For dense predictions in the segmentation task, the upsampling operation following by a softmax layer is used in the mask predictor $P$ for probability predictions of the pixels.

To transfer the label information from the source domain to the target domain, supervised learning with the source domain $\mathcal{D}_s$ is an essential part of domain adaptation. The $\mathcal{L}_{seg}$ loss is defined as follows
\begin{equation}\label{eq:segloss}
\small
\begin{split}
\mathcal{L}_{seg} 
	= & -\frac{1}{HW}\sum \limits_{h=1}^{H} \sum \limits_{w=1}^{W} \sum \limits_{c=1}^{N_c}y^{(h,w,c)}_s \log(M^{(h,w,c)}_s) \\
& -\eta \sum \limits_{c=1}^{N_c} \frac{ \sum \limits_{h=1}^{H} \sum \limits_{w=1}^{W} 2 y^{(h,w,c)}_s M^{(h,w,c)}_s }{  \sum \limits_{h=1}^{H} \sum \limits_{w=1}^{W} y^{(h,w,c)}_s + \sum \limits_{h=1}^{H} \sum \limits_{w=1}^{W}  M^{(h,w,c)}_s }\,,
\end{split}
\end{equation}
where $\mathbf{y}_s \in \mathbb{R}^{ H\times W \times N_c } $ denotes one-hot labels, and $\eta \in [0, 1]$ is a trade-off parameter. The first term is the cross-entropy loss for pixelwise classification. The second term is the Dice loss for multiple cardiac structures, which is commonly employed in biomedical image segmentation tasks. We combine the two complementary hybrid loss functions to address the challenging heart segmentation task.

\begin{figure*}[htb]
\centering
\includegraphics[width=0.99\textwidth]{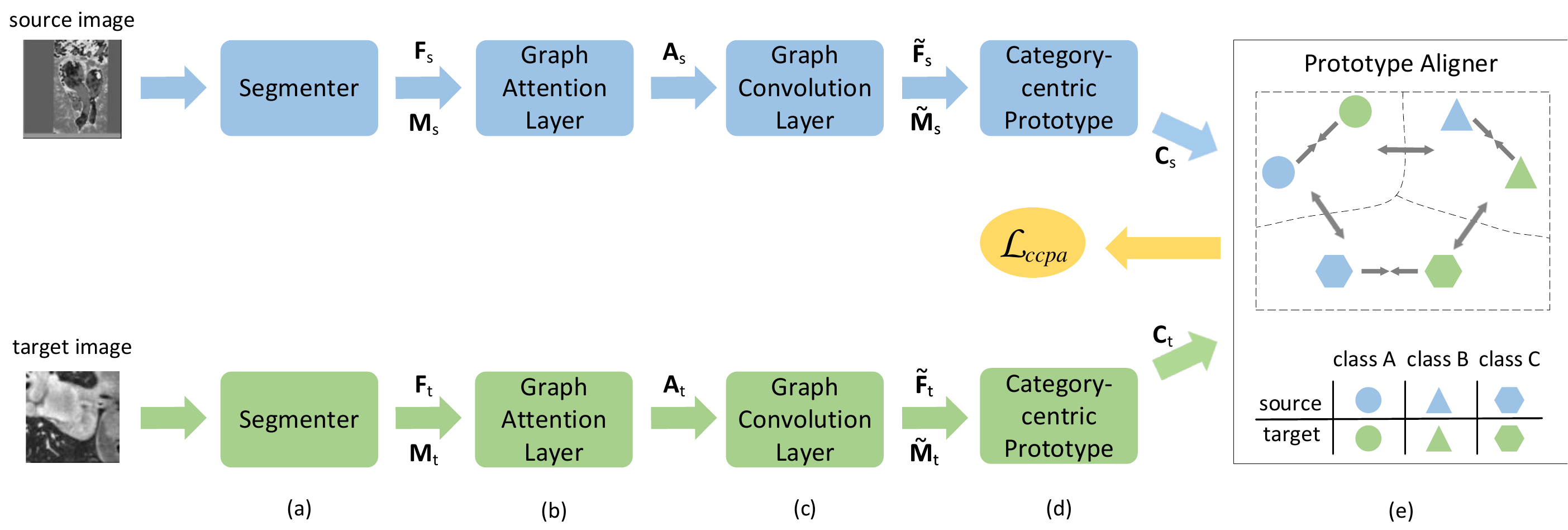}
\caption{Category-Centric Prototype Aligner flowchart.  (a)~Feature representation $\mathbf{F}$ and segmentation prediction $\mathbf{M}$  are generated via the segmenter.  (b)~Pixel-level relational weights $\mathbf{A}$ are obtained by the graph attention layer. (c)~More accurate instance-level feature representations $\widetilde{\mathbf{F}}$ and $\widetilde{\mathbf{M}}$ are acquired through the graph convolution layer. (d)~Category-centric prototype $\mathbf{C}$ is derived via confidence-guided merging. (e)~Performing category-level domain alignment by minimizing loss $\mathcal{L}_{ccpa}$ calculated by prototype aligner.}
\label{fig:ccpa}
\end{figure*}

\subsection{Conditional Domain Discriminator}
\label{sec:cdd}
Existing UDA works~\cite{Dou2018UnsupervisedCD,Dou2019PnPAdaNetPA} use dual-domain discriminators whose inputs are the feature-level features and the predicted segmentation masks of the source and target domains respectively. These works aim to align the latent feature space of the target domain to that of the source domain, with a more explicit constraint on the shape of segmentation masks. Although the starting point for these works~\cite{Dou2018UnsupervisedCD,Dou2019PnPAdaNetPA} is good and performs well on datasets, how to balance the ratio between dual discriminators can severely affect the domain adaptation performance, as it needs empirical design and time for UDA. Additionally, due to the lack of the supervisory signal of the target domain, the predicted map $\mathbf{M}_t$ probably have error noise or inaccurate boundaries, which makes it harder to align the source domain and target domain on the instance level when data distributions embody complex multimodal structures.

In this regard, we incorporate CDD inspired by \cite{Long2018ConditionalAD} into the existing architecture illustrated in the left-middle position of Figure~\ref{fig:overall} to effectively align different domains of multimodal distributions native in segmentation problems. This is different from aligning the features and segmentation maps separately~\cite{Tsai2018LearningTA,Chen2017NoMD}. Notably, this module can capture the cross-covariance between feature representations and segmentation predictions to improve the transferability. The key to CDD models is a novel conditional domain discriminator conditioned on the cross-covariance of domain-specific feature representations and segmentation predictions. 

Figure~\ref{fig:discriminator}(c) shows a detailed description of the CDD. Given two inputs $\mathbf{F}$ and $\mathbf{M}$ of the source domain and target domain respectively, the CDD models the multimodal information and joint distributions of $\mathbf{F}$ and $\mathbf{M}$ via a randomized router, which makes it easier to fully capture multiplicative interactions between feature representation and segmentation prediction. Then, conditional domain adversarial loss measures the domain discrepancy by training a domain discriminator in a conditional manner. Given conditional domain discriminator $D_{cdd}$, the definition of conditional domain adversarial loss is
\begin{equation}\label{eq:cddloss}
\small
\begin{split}
\mathcal{L}_{cdd}(\mathcal{D}_s, \mathcal{D}_t)
	= & \mathbb{E}_{x^s\sim \mathcal{D}_s}\log[D_{cdd}(J(\mathbf{F}_s, \mathbf{M}_s))] \\
& + \mathbb{E}_{x^t\sim \mathcal{D}_t}\log[D_{cdd}(J(\mathbf{F}_t, \mathbf{M}_t))]\,,
\end{split}
\end{equation} 
where $J$ is the explicit randomized multilinear map that converts $\mathbf{F}$ and $\mathbf{M}$ to a single tensor that is the joint variable of domain-specific feature representation and segmentation prediction for adversarial adaptation. We define
\begin{equation}
J(\mathbf{F}, \mathbf{M})
 = \frac{1}{{\sqrt{d_o} }}\left( {{{\mathbf{F}} {\mathbf{R}}_{\mathbf{F}}}} \right) \odot \left( {{\mathbf{M}} {{\mathbf{R}}_{\mathbf{M}}}} \right) \in \mathbb{R}^{ H\times W \times d_o } \,,
\end{equation} 
where $\odot$ represents an element-wise product, $d_o$ is the dimension of the output tensor, and ${\mathbf{ R}_\mathbf{F}} \in \mathbb{R}^{ d \times d_o}$ and ${\mathbf{ R}_\mathbf{M} \in \mathbb{R}^{ N_c \times d_o} }$ are random matrices, in which each element $R_{ij}$ follows a symmetric distribution, where $\mathbb{E}\left[ {R_{ij} } \right] = 0$ and $\mathbb{E}[ {{{{R_{ij}^2 }}}} ] = 1$. In practice, $\mathbf{ R}_\mathbf{F} $ and $ \mathbf{ R}_\mathbf{M}$ are sampled only once from a uniform distribution or Gaussian distribution and fixed in the training phase. We restrict $d_o \ll d \times N_c$ of random matrices to avoid dimension explosion~\cite{Kar2012RandomFM,Rahimi2007RandomFF, Long2018ConditionalAD}.

One common optimization approach for adversarial learning networks in an unsupervised manner is to follow the training rules of GAN~\cite{Arjovsky2017WassersteinG}, which is implemented in the form of a minimax two-player game containing a generator and a discriminator role. An alternative optimization method utilizes a gradient reversal layer~(GRL) ~\cite{Ganin2015UnsupervisedDA}, which is inserted between the feature generator and CDD. In practice, we use the latter implementation method to align the domain statistics with CDD that predicts the domain.

\subsection{Category-Centric Prototype Aligner}
\label{sec:ccpa}
The gap between the two domains can severely decrease the model's performance. Furthermore, as different instances commonly embody distinct modal information in the cross-modality scenario, especially when neighboring structures remain unclear boundaries or relatively homogeneous tissues\cite{Dou2019PnPAdaNetPA}, combining the feature-level and instance-level alignment of the source and target domains makes it difficult to perfectly address domain shift~\cite{Xu2020CrossDomainDV, Pan2019TransferrablePN, Saito2019SemiSupervisedDA, Xie2018LearningSR}. In this regard, in addition to the CDD module, we also proposed CCPA shown in the right-middle position of Figure~\ref{fig:overall}, inspired by~\cite{Xu2020CrossDomainDV}, which includes five steps to align the source and target domains with category-centric prototype representations. The detailed flowchart is illustrated in Figure~\ref{fig:ccpa}. In addition, to relieve the negative effect of class imbalance~\cite{Lin2020FocalLF} on domain adaptation, we design a category-reweighted contrastive loss to balance the training process of domain adaptation.

\noindent \textbf{Graph Attention Layer} As described in Section~\ref{sec:segmenter}, feature representation $\mathbf{F}$ and segmentation prediction $\mathbf{M}$ are generated via a segmenter on both the source and target domains, which is the first step of the flowchart shown in Figure~\ref{fig:ccpa}(a). Then, we use the graph attention layer~\cite{Velickovic2018GraphAN, Vaswani2017AttentionIA} to produce adjacency matrix $\mathbf{A} \in \mathbb{R}^{HW \times HW}$, which is used to model the relationship between pixels. Intuitively, two spatially closer pixels are more likely to depict the same class and should be assigned higher connection weights. Following this idea, a method for obtaining an adjacency matrix is defined as
\begin{small}
\begin{equation}
\begin{cases}
\textbf{A}_{ij} = \mathrm{softmax}_j (\mathbf{e}_{ij})\,, \quad i\,, j \in [1, HW] \,,\\
\mathrm{\mathbf{e}}_{ij} = \mathrm{LeakRelu}(\mathbf{w}_i^T| \bm{h}_i -  \bm{h}_j|))\,,
\end{cases}
\end{equation}
\end{small}where $\mathbf{w}_i \in \mathbb{R}^d$ is the learnable weight vector, $\mathbf{h}_i \in \mathbb{R}^d$ and $\mathbf{F} = \{\mathbf{h}_i\}_1^{HW}$. To solve the problem of gradients vanishing in the training phase, we use LeakyReLU instead of the Relu activation function. In practice, to avoid high computing complexity and memory, we perform graph attention on downsample feature representation $\mathbf{F}$ and then upsample matrix $\mathbf{A}$ to the original resolution size.


\noindent \textbf{Graph Convolution Layer} Because of the boundary deviation in segmentation prediction, often distributed around the ground truth mask, initial segmentation prediction conveys incomplete instance information, which leads to inaccurately of representing an instance. A natural approach here is that the segmentation prediction feature belonging to a certain instance should be aggregated, to achieve exact instance-level feature representations. Specifically, more exact instance-level feature representations are calculated by using an adjacency matrix $\mathbf{A}$ containing the spatial relevance,  image features $\mathbf{F}$ and segmentation confidence $\mathbf{M} $. Formulation is defined as
\begin{equation}\label{eq:aggeratef}
\widetilde{\mathbf{F}} =\sigma( \mathbf{A} \mathbf{F}\mathbf{W}_F) \in \mathbb{R}^{HW \times d}\,,
\end{equation}
\begin{equation}\label{eq:aggeratem}
\widetilde{\mathbf{M}} =\sigma( \mathbf{A} \mathbf{M}\mathbf{W}_M) \in \mathbb{R}^{HW \times N_c}\,,
\end{equation}
where $\mathbf{W}_F \in \mathbb{R}^{d \times d}$ and $\mathbf{W}_M \in \mathbb{R}^{N_c \times N_c}$ are learnable weight matrices. $\sigma(\cdot)=\max(0,\cdot)$ is a nonlinear activation function. In Eq. \ref{eq:aggeratef}, \ref{eq:aggeratem}, after graph convolution, $\widetilde{\mathbf{F}}$ and $\widetilde{\mathbf{M}}$ more precise instance-level information is aggregated through information propagation among adjacent pixels.


\noindent \textbf{Category-centric Prototype} After obtaining more accurate feature representations aggregated on the instance level, we employ confidence-guided metering to integrate the multimodal information reflected by different instances into prototype representations. Category-centric prototypes $\mathbf{C} \in \mathbb{R}^{N_c \times d}$ are calculated by the weighted mean feature representation $\widetilde{\mathbf{F}}$:
\begin{small}
\begin{equation}
\begin{cases}
\mathbf{C} = \boldsymbol{\alpha}^T \widetilde{\mathbf{F}} \in \mathbb{R}^{N_c \times d} \,,\\
\alpha_{i,j}= \frac{\widetilde{\textbf{P}}_{i,j}}{\sum_{i=1}^{HW} \, \widetilde{\textbf{P}}_{i,j}} \,,
\end{cases}
\end{equation}
\end{small}where $\boldsymbol{\alpha} \in \mathbb{R}^{HW \times N_c}$ represents normalized segmentation confidence to each class. The derived prototypes $\mathbf{C}$ serve as the proxy of each class during subsequent domain alignment. 


\noindent \textbf{Prototype Aligner} Following the heuristic rules of the prototype-based method~\cite{Saito2019SemiSupervisedDA, Xu2020CrossDomainDV} for unsupervised domain alignment, we minimize intraclass loss to narrow the distance between the same categories' prototypes of two domains, named $\mathcal{L}_{intra}$. In addition, we also minimize interclass loss to bound the distance between different categories' prototypes, named $\mathcal{L}_{inter}$. Furthermore, in segmentation tasks, as a class-imbalance problem usually exists~\cite{Lin2020FocalLF}, we address this problem through reweight loss $\mathcal{L}_{intra}$ and $\mathcal{L}_{inter}$ using an entropy-based map~\cite{Vu2019ADVENTAE}. The underlying idea is that hard samples or sample-scarce categories produce high-entropy predictions on both source and target domains because the categories with abundant samples are trained more sufficiently and better aligned so that they have higher confidence compared with sample-scarce categories, and vice versa. 

In this regard, we assign higher weights to the sample-scarce categories during the training process of domain adaptation. The weight corresponding to the $k$-th category is calculated by:
\begin{equation}\label{eq:prototypeweight}
\small
\beta_k
	=  -\frac{1}{HW}\sum \limits_{h=1}^{H} \sum \limits_{w=1}^{W} M^{(h,w,k)} \log(M^{(h,w,k)}) \,.
\end{equation}

Note that the intraclass loss requires prototypes of the same category to be as close as possible, and the interclass loss constrains the distance between prototypes of different classes to be larger than a margin. The category-centric prototype domain adaptation loss $\mathcal{L}_{ccpa}$ consists of an intraclass loss and three interclass losses in which 
all pairwise relations between two domains' prototypes are considered. So we define the following loss:
\begin{small}
	\begin{equation}\label{eq:ccpaloss}
	\begin{split}
	\mathcal{L}_{ccpa} = \ \mathcal{L}_{intra} & (\mathcal{D}_s, \mathcal{D}_t) + \frac{1}{3} \, \big( \mathcal{L}_{inter} (\mathcal{D}_s, \mathcal{D}_s) \\
	& + \mathcal{L}_{inter} (\mathcal{D}_s, \mathcal{D}_t) + \mathcal{L}_{inter} (\mathcal{D}_t, \mathcal{D}_t) \big) \,, 
	\end{split}
	\end{equation}
	\begin{equation} \label{eq:intraloss}
	\mathcal{L}_{intra} (\mathcal{D}_s, \mathcal{D}_t) = \frac{\sum_{i=1}^{N_c} \, \beta_i^{s} \beta_i^{t}\Phi(C_i^s, C_i^t)}{\sum_{i=1}^{N_c} \, \beta_i^s \beta_i^t} \,,
	\end{equation}
	\begin{equation} \label{eq:interloss}
	\mathcal{L}_{inter} (\mathcal{D}, \mathcal{D}') = \frac{\sum \limits_{1 \leqslant i \neq j \leqslant N_c} \! \! \beta_i^{\mathcal{D}} \beta_j^{\mathcal{D}'} max(0, m - \Phi(C_i^{\mathcal{D}}, C_j^{\mathcal{D}'}))}{\sum \limits_{1 \leqslant i \neq j \leqslant N_c} \! \! \beta_i^{\mathcal{D}} \beta_j^{\mathcal{D}'}} \,,
	\end{equation}
\end{small}where $\Phi(c,c') = ||c - c'||_2$ is the Euclidean distance between two prototypes and $\{C_i^s\}_{i=1}^{N_c}$, $\{C_i^t\}_{i=1}^{N_c}$ represent the prototypes of source and target domains. $\mathcal{D}$ and $\mathcal{D}'$ denote two domains from which prototype pairs belonging to different categories are drawn. $m$ is the margin term which is fixed as $1.0$ in all experiments.

Our model parameters of whole networks are jointly trained by minimizing the following total loss: 
\begin{equation}\label{eq:totalloss}
\mathcal{L}_{total} = \mathcal{L}_{seg} + \lambda_1 \mathcal{L}_{cdd}  + \lambda_2 \mathcal{L}_{ccpa} \,,
\end{equation}
where $\mathcal{L}_{seg}$, $\mathcal{L}_{cdd}$, and $\mathcal{L}_{ccpa}$ are defined in Eqs.~\ref{eq:segloss}, ~\ref{eq:cddloss}, and~\ref{eq:ccpaloss}, respectively. $\lambda_1$ and $\lambda_2 \in [0, 1]$ are trade-off parameters.

\section{Experiments}

\subsection{Experimental Details}
\textbf{Datasets} We use a medical cross-modality domain adaptation benchmark proposed in~\cite{Dou2019PnPAdaNetPA} to validate the performance of our proposed unsupervised cross-modality domain adaptation method for biomedical image segmentation. This dataset contains training (16 subjects) and testing (4 subjects) sets for each modality, and it collects from the public dataset of \textit{MICCAI 2017 Multi-Modality Whole-Heart Segmentation}~\cite{Zhuang2016MultiscalePA}, which consists of 20 unpaired CT and 20 MRI images from 40 patients. The CT and MRI images were obtained in different clinical centers. The cardiac structures of the images were manually annotated by radiologists for both MRI and CT images. Our segmenter aimed to automatically segment four cardiac structures, including the ascending aorta (AA), the left atrium blood cavity (LA-blood), the left ventricle blood cavity (LV-blood), and the myocardium of the left ventricle (LV-myo). All the volumetric MRI and CT images and corresponding labels were preprocessed in~\cite{Dou2018UnsupervisedCD, Dou2019PnPAdaNetPA}.

\begin{table*}[t]
\caption{Quantitative performance comparison between different methods on cardiac datasets~ (MRI $\rightarrow$ CT). (Note: \textbf{Bold} represent the best performance and the - denotes that the results were not reported by that method.)}
\centering
\small
\renewcommand{\arraystretch}{1.2}
\begin{center}
\begin{tabular}{ |m{3.46cm}|m{1.3cm}|m{1.3cm}|m{1.3cm}|m{1.3cm}|m{1.3cm}|m{1.3cm}|m{1.3cm}|m{1.3cm}|  }
\hline
\multirow{2}[1]{*}{\centering Methods }  & \multicolumn{2}{c|}{AA} & \multicolumn{2}{c|}{LA-blood} & \multicolumn{2}{c|}{LV-blood} & \multicolumn{2}{c|}{LV-myo} \\
\cline{2-9}
&~~~ Dice & ~~~ ASD & ~~~ Dice & ~~~ ASD & ~~~ Dice & ~~~ ASD & ~~~ Dice  & ~~~ ASD \\
\hline

U-Net-MRI~\cite{Ronneberger2015UNetCN}  & 70.6$\pm$20.2 & ~~~~~~~  - & 74.0$\pm$24.7 & ~~~~~~~  -  & 81.1$\pm$23.8 & ~~~~~~~  -  & 68.1$\pm$25.3 & ~~~~~~~  - \\

Cascaded-FCN-MRI~\cite{Payer2017MultilabelWH}  & 76.6$\pm$13.8 & ~~~~~~~  - & 81.1$\pm$13.8 & ~~~~~~~  -  & 87.7$\pm$7.7 & ~~~~~~~  -  & 75.2$\pm$12.1 & ~~~~~~~  - \\


PnP-AdaNet-v2-MRI~\cite{Dou2019PnPAdaNetPA}    & 80.3$\pm$1.8 &  ~~~~~~~  -  & 78.1$\pm$8.0 &  ~~~~~~~  -  & 88.3$\pm$2.1 &  ~~~~~~~  -   & 70.8$\pm$2.1 &  ~~~~~~~  - \\

U-Net-CT~\cite{Ronneberger2015UNetCN}  & 94.0$\pm$6.2 & ~~~~~~~  - & 91.0$\pm$5.2 & ~~~~~~~  -  & 91.0$\pm$4.3 & ~~~~~~~  -  & 86.1$\pm$4.2 & ~~~~~~~  - \\

Cascaded-FCN-CT~\cite{Payer2017MultilabelWH}      & 91.1$\pm$18.4 &~~~~~~~   - & 92.4$\pm$3.6 & ~~~~~~~  -  & 92.4$\pm$3.3 &~~~~~~~   -  & 87.2$\pm$3.9 & ~~~~~~~  - \\


PnP-AdaNet-v2-CT~\cite{Dou2019PnPAdaNetPA}    & 77.6$\pm$29.2  &  ~~~~~~~  -  & 89.4$\pm$2.2 &  ~~~~~~~  -   & 91.3$\pm$2.5 &  ~~~~~~~  -  & 85.8$\pm$1.5 &  ~~~~~~~  - \\

\hline
Seg-MRI~(ours)   & 78.3$\pm$5.5  & 15.3$\pm$10.6  & 78.9$\pm$7.1 & 15.7$\pm$6.4   & 88.5$\pm$5.2 & ~~2.9$\pm$0.5   & 75.3$\pm$3.1 & ~~2.3$\pm$2.1 \\

Seg-CT~(ours)      & 84.9$\pm$21.4 & ~~2.0$\pm$9.7  & 88.3$\pm$6.1 & ~5.8$\pm$7.4   & 91.5$\pm$4.6 & ~~5.3 $\pm$6.1  & 86.2$\pm$9.7 & ~~5.9$\pm$11.4\\
\hline
\hline


Seg-CT-noDA ~(ours) & 32.3$\pm$8.4  & 14.4$\pm$19.3 & 28.9$\pm$14.3& ~~8.9$\pm$3.8  & ~~2.9$\pm$1.1  & 19.2 $\pm$9.4 & 15.7$\pm$13.1 & 28.4$\pm$33.9\\

\hline
\hline

		DANN-CT-DA~\cite{Ganin2016DomainAdversarialTO}        & 39.0$\pm$35.1 & 16.2$\pm$5.8 & 45.1$\pm$23.6 & 9.2$\pm$2.9  & 28.3$\pm$11.8 & 12.1$\pm$1.6  & 25.7$\pm$13.2 & 10.1$\pm$3.2 \\
		ADDA-CT-DA~\cite{Tzeng2017AdversarialDD}   & 47.6$\pm$15.2 & 13.8$\pm$3.0 & 60.9$\pm$13.2 & 10.2$\pm$6.5 & 11.2$\pm$13.1 &  ~~~~~~~  -  & 29.2$\pm$16.4 & 13.4$\pm$5.0 \\
		CycleGAN-CT-DA~\cite{Zhu2017UnpairedIT}       & 73.8$\pm$7.4  & 11.5$\pm$2.9 & 75.7$\pm$4.3 & 13.6$\pm$3.6 & 52.3$\pm$21.0 & 9.2$\pm$3.9 & 28.7$\pm$13.3 & ~8.8$\pm$4.3 \\
		PnP-AdaNet-v2-CT-DA  ~\cite{Dou2019PnPAdaNetPA}             & 74.0$\pm$7.3  & 12.8$\pm$3.2  & 68.9$\pm$5.2 & 6.3$\pm$2.3  & 61.9$\pm$10.7 & 17.4$\pm$7.0 & 50.8$\pm$7.0  & 14.7$\pm$4.8\\
		
			
			SIFA-v2-CT-DA  ~\cite{Chen2020UnsupervisedBC}             &~~~~~81.3 &~~~~~7.9  &~~~~~\textbf{79.5} &  ~~~~~6.2 & ~~~~~73.8  & ~~~~~\textbf{5.5}  & ~~~~~61.6  & ~~~~~8.5 \\

	3D based-CT-DA  ~\cite{Ouyang2019DataEU}             & 78.3$\pm$13.2  & \textbf{2.7$\pm$3.2}  & 78.2$\pm$9.8 & \textbf{3.8$\pm$1.2}  & 71.8$\pm$17.9 & 6.4$\pm$2.1 & 60.2$\pm$15.8  & 7.3$\pm$10.8\\
				
\hline
Seg-CT-DA~(ours) &\textbf{ 81.8$\pm$18.7} & 8.1$\pm$7.9 & 76.9$\pm$22.6 & 7.6$\pm$8.5 & \textbf{75.9$\pm$15.8} & 7.5$\pm$6.1 & \textbf{62.3$\pm$16.1} &\textbf{7.1$\pm$6.9}\\

\hline
\end{tabular}
\end{center}
\label{tab:expresults}
\end{table*}

\noindent\textbf{Evaluation metrics} We employed the Dice coefficient $\!([\%])\!$ to evaluate the agreement between the ground truth and predicted segmentation for cardiac structures. In addition, we calculated the average surface distance (ASD$[\text{voxel}]$) to measure the segmentation performance from the perspective of the boundary. A higher Dice and lower ASD reflect better segmentation performance. Both metrics are presented in the format of \emph{mean$\pm$std}, which shows the average performance as well as the cross-subject variations of the results.

\noindent\textbf{Network architectures} In the segmenter, we use DRN101~\cite{Yu2017DilatedRN} as the base semantic segmentation architecture, which {is composed of stacked dilated residual blocks, to capture a large receptive field and the spatial context of feature maps. The DRN hyper-parameters used in our paper are the same as~\cite{Dou2019PnPAdaNetPA}. The multiple stages of layers [1, 3, 5, 7] are concatenated as image features $\mathbf{F}$. The image feature dimension is $d=512$. We modify the stride and dilation rate of the last layers followed by an upsampling layer to produce denser feature maps with a larger field of view for segmentation. The CDD consists of several stacked residual blocks followed by a sigmoid layer to predict the domain. The output dimension of the randomized multilinear map $d_o=128$.
\\
\textbf{Training details} The proposed model is implemented in PyTorch and trained on 4 NVIDIA RTX 2080 Ti GPUs with 44 GB memory. We train using the Adam optimizer [23] in three stages, and the batch size is 16, split equally for source and target samples. First, our model is trained from scratch on the source domain using a learning rate of $1e-3$ to minimize the segmentation loss $\mathcal{L}_{seg}$. The trade-off parameter $\eta$ in the $\mathcal{L}_{seg}$~(Eq.~\ref{eq:segloss}) item is 1. Second, the network is trained with only the source supervision and unsupervised domain classification losses $\mathcal{L}_{seg} + \lambda_1 \mathcal{L}_{cdd}$ at a learning rate of $3e-4$. Then, the overall loss function~(Eq.~\ref{eq:totalloss}) is optimized, applying the conditional domain adversarial loss $\mathcal{L}_{cdd} $ and category-centric prototype alignment loss $\mathcal{L}_{ccpa} $ jointly, and the learning rate is set to $1e-4$. The trade-off ratio parameters of  $\lambda_1$ and $\lambda_2$ are both set to the default standard setting value of 1 in the loss $\mathcal{L}_{ccpa}$. We also perform a series of experiments to analyze the impact of different hyper-parameters including $\lambda_1, \lambda_2$ and $\eta$, in the ablation study section. We use dropout with a ratio of 0.1 and batch normalization in all the convolutional layers. At the inference phase, the model directly predicts every pixel that belongs to the most likely class via the segmenter on the source or target domain.
\\
\textbf{Experimental settings} To verify the performance of our proposed method, we conducted extensive experiments to demonstrate the performance of the domain adaptation method. In our experiments, biomedical cardiac MRI images were the source domain, and CT images were the target domain. In practice, we carry out the following experimental settings:\\
1) Training and testing the segmentation network only on the source domain (referred to Seg-MRI);\\
2) Training on the source domain and directly testing on target data, with no domain adaptation, as a lower bound (referred to Seg-CT-noDA); \\
3) Training and testing the segmentation network on annotated target domain images, as an upper bound (referred to  Seg-CT); \\
4) Our method for unsupervised domain adaptation (referred to Seg-CT-DA).

\begin{table*}[t]
\caption{Performance comparisons of the different components of our model on cardiac datasets~(MRI $\rightarrow$ CT). The results indicate the importance of the proposed module for unsupervised domain adaptation. (Note: \textbf{Bold} represent the best performance.)}
\centering
\small
\renewcommand{\arraystretch}{1.2}
\begin{center}
\begin{tabular}{ |m{3.46cm}|m{1.3cm}|m{1.3cm}|m{1.3cm}|m{1.3cm}|m{1.3cm}|m{1.3cm}|m{1.3cm}|m{1.3cm}|  }
\hline
\multirow{2}[1]{*}{\centering Methods }  & \multicolumn{2}{c|}{AA} & \multicolumn{2}{c|}{LA-blood} & \multicolumn{2}{c|}{LV-blood} & \multicolumn{2}{c|}{LV-myo} \\
\cline{2-9}
&~~~ Dice & ~~~ ASD & ~~~ Dice & ~~~ ASD & ~~~ Dice & ~~~ ASD & ~~~ Dice  & ~~~ ASD \\

\hline
Seg-CT-DA~(Standard Setting: $\lambda_1=1, \lambda_2=1, \eta=1$) & \textbf{81.8$\pm$18.7} & ~~\textbf{8.1$\pm$7.9} & \textbf{76.9$\pm$22.6} & ~~\textbf{7.6$\pm$8.5} & ~\textbf{75.9$\pm$15.8} & \textbf{7.5$\pm$6.1} & \textbf{62.3$\pm$16.1} & \textbf{7.1$\pm$6.9}\\

\hline
\hline

w/o CDD~($\lambda_1=0$)     & 73.6$\pm$16.1 & 11.7$\pm$14.4  & 68.2$\pm$9.2 & ~~8.1$\pm$6.1   & ~~60.1$\pm$8.6 & ~9.8 $\pm$7.5  & 49.6$\pm$7.4 & 15.4$\pm$15.3\\

w/o CCPA~($\lambda_2=0$)  &74.9$\pm$13.7  & 12.5$\pm$14.3 & 66.2$\pm$18.6 & ~~7.9$\pm$4.3  & ~~57.8$\pm$3.5  & ~9.5 $\pm$9.6 & 48.2$\pm$14.7 & 14.3$\pm$24.7\\

w/o CDD \& CCPA  & 32.3$\pm$8.4  & 14.4$\pm$19.3 & 28.9$\pm$14.3& ~~8.9$\pm$3.8  & ~~2.9$\pm$1.1  & 19.2 $\pm$9.4 & 15.7$\pm$13.1 & 28.4$\pm$33.9\\

\hline
\hline

		$\lambda_1=0.3$      & 75.4$\pm$31.5 & 10.4$\pm$8.9 & 71.9$\pm$12.8 & ~~8.0$\pm$5.6 & 64.8$\pm$12.5 &  ~~8.6$\pm$5.3  & 53.8$\pm$30.2 & 13.1$\pm$3.7 \\
		$\lambda_1=0.6$       & 78.5$\pm$28.2  & ~9.7$\pm$8.3 & 75.3$\pm$3.9 & ~~7.8$\pm$4.3 & 68.3$\pm$32.0 & ~~8.0$\pm$9.3 & 56.9$\pm$21.5 & ~10.4$\pm$3.3 \\
		$\lambda_1=0.9$      & 80.3$\pm$31.8 & ~8.8$\pm$5.3 & 76.1$\pm$10.2 & ~~7.7$\pm$5.2 & 73.2$\pm$21.7 & ~~7.8$\pm$7.1 & 61.4$\pm$22.7 & ~~8.2$\pm$7.2  \\ 

\hline
\hline
		
		 $\lambda_2=0.3$           & 77.3$\pm$12.5 & 10.2$\pm$4.2  & 70.5$\pm$5.3 & ~~7.9$\pm$2.6  & 65.3$\pm$22.5& ~~8.9$\pm$6.2 & 52.5$\pm$31.7& ~~12.5$\pm$4.2\\
			
		 $\lambda_2=0.6$              &79.6$\pm$22.7 & ~9.4$\pm$5.4  & 72.3$\pm$2.5 &  ~~7.8$\pm$3.5 & 70.8$\pm$11.8  & ~~8.4$\pm$2.5  & 56.2$\pm$22.8  & ~~9.3$\pm$2.4 \\

	     $\lambda_2=0.9$            & 81.1$\pm$6.4  &  ~8.6$\pm$7.1  & 75.1$\pm$4.9 & ~~7.7$\pm$4.8 & 74.3$\pm$26.4 &    ~~7.6$\pm$2.8 & 61.9$\pm$18.1  & ~~7.8$\pm$3.7\\
		
\hline
\hline

		 $\eta=0$          & 76.0$\pm$18.2  & ~9.6$\pm$5.7  & 70.1$\pm$1.9 & ~~8.5$\pm$3.2  & 67.2$\pm$12.4 & ~~8.7$\pm$4.9 & 54.8$\pm$12.7  & 11.7$\pm$8.2\\
		
		 $\eta=0.2$          &78.3$\pm$24.6 &  ~9.2$\pm$3.3 & 71.5$\pm$3.5& ~~8.2$\pm$4.5  & 69.6$\pm$23.5 & ~~8.4$\pm$2.4 &56.7$\pm$13.6 & 10.1$\pm$3.0\\
			
		 $\eta=0.4$              &79.6$\pm$14.9 & ~8.8$\pm$6.7   & 73.4$\pm$8.4 & ~~8.0$\pm$5.4 & 71.5$\pm$31.6   & ~~8.0$\pm$4.1  & 58.1$\pm$17.6   & ~9.4$\pm$3.6 \\

	 $\eta=0.6$            & 80.6$\pm$6.8  & ~8.6$\pm$2.4 & 74.2$\pm$4.6 & ~~7.9$\pm$4.1 & 72.8$\pm$21.8 & ~~7.8$\pm$2.9 & 60.8$\pm$11.9  & ~8.6$\pm$4.1\\
	 
	 	 $\eta=0.8$            & 81.4$\pm$24.7  & ~8.3$\pm$2.1  & 75.6$\pm$8.9 & ~~7.7$\pm$2.9  & 74.0$\pm$14.3 & ~~7.7$\pm$6.4 & 61.3$\pm$17.5  & ~7.4$\pm$5.3\\
	 	 		
\hline
\end{tabular}
\end{center}
\label{tab:ablation}
\end{table*}

\begin{figure*}[htb]
\centering
\includegraphics[width=0.99\textwidth]{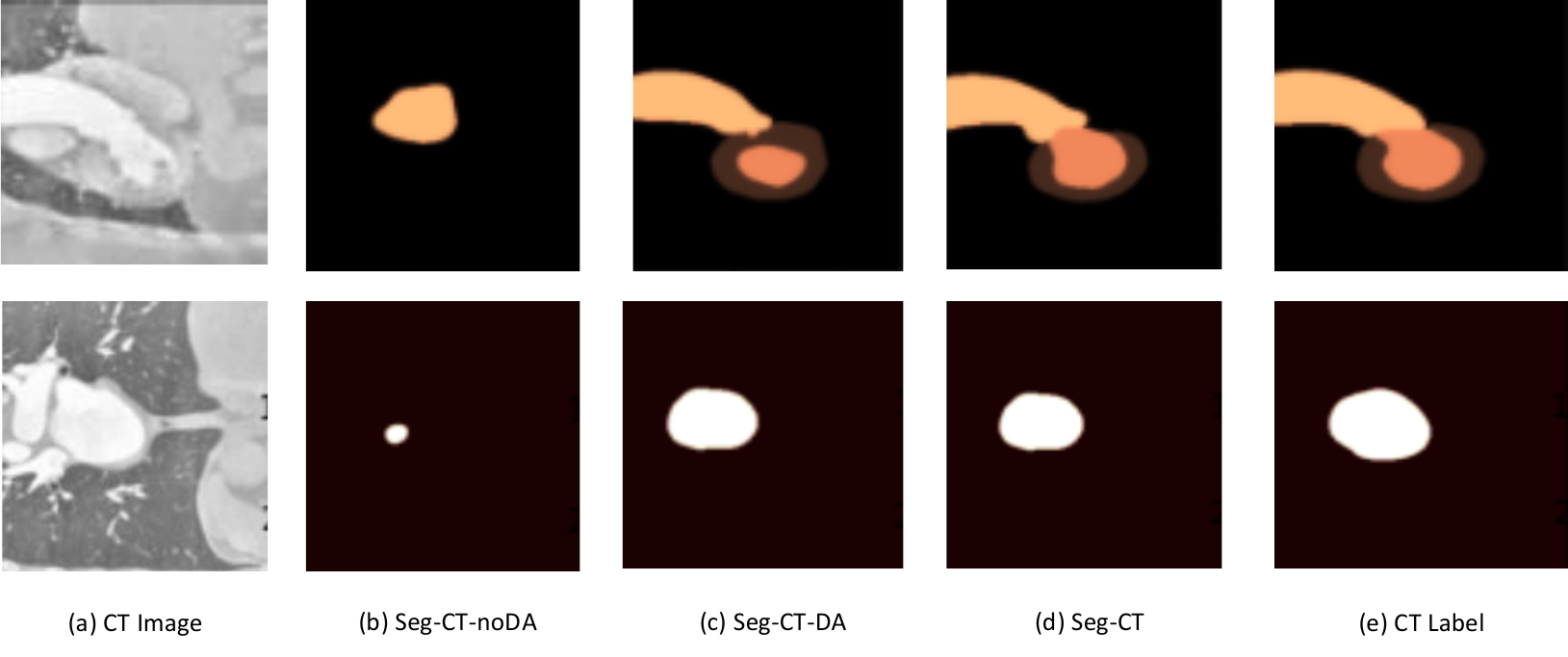}
\caption{Results of different methods for CT image segmentations. Each row presents one typical example, from left to right: (a) raw CT image, (b) directly applying the MRI segmenter on CT data, (c) our unsupervised cross-modality domain adaptation result, (d) the segmenter trained from scratch with CT labels, and (e) ground truth labels. The structures of AA, LA-blood, LV-blood and LV-myo are indicated by brown, white, red and light red colors, respectively.}
\label{fig:resultsvis}
\end{figure*}

\subsection{Experimental Results}
We report segmentation quantitative results in this section, which demonstrate the superiority of the proposed method in the UDA scenario for cardiac structure segmentation, as illustrated in Table~\ref{tab:expresults}. In addition, Figure~\ref{fig:resultsvis} presents the qualitative results of the segmentation for CT images.

First, we validated the performance of the segmenter for Seg-MRI, which serves as the basis for subsequent domain adaptation procedures. Our segmenter achieved competitive performance on most of the four cardiac structures compared with the standard U-Net~\cite{Ronneberger2015UNetCN} and cascaded-FCN~\cite{Payer2017MultilabelWH} methods. With the segmenter network architecture, we performed the following experiments to validate the effectiveness of our unsupervised domain adaptation framework.

Then we confirmed the upper-bounds performance of the segmenter on the target domain and report it as Seg-CT. Generally, these results are comparable to the standard U-Net~\cite{Ronneberger2015UNetCN} and cascaded-FCN~\cite{Payer2017MultilabelWH} methods. Furthermore, comparing Seg-MRI with Seg-CT, we found a significant performance gap, which demonstrates the severe domain shift between the source and target domains. Furthermore, domain-shift problem inherent in cross-modality biomedical images is also illustrated by the degradation of Seg-CT-noDA's performance. This indicates that although the cardiac MRI and CT images share similar high-level representations and identical label space, the geometric pattern or boundary of each category or instance remains different, which makes domain adaptation extremely difficult.

What is striking about the figures in this table is that Seg-CT-DA outperforms Seg-CT-noDA in all metrics. Further analysis showed that the most striking aspect of the data was the largest increase in Dice of LV-blood. These results show the benefits of using a category-centric prototype to distinguish modal information in the cross-modality scenario, especially when neighboring structures contain unclear boundaries or relatively homogeneous tissues. Furthermore, as seen in the last part in Table~\ref{tab:expresults}, our method shows significant improvement over other domain adaptation methods in half of the metric. These results demonstrate the superiority of the proposed method in the UDA scenario for cardiac structure segmentation. 
In summary, these results show the importance of using CDD and CCPA across domains.

\begin{table*}[t]
\caption{Quantitative performance  of reverting the domain adaptation direction comparison between different methods on cardiac datasets~(CT $\rightarrow$ MRI). (Note: \textbf{Bold} represent the best performance.) }
\centering
\small
\renewcommand{\arraystretch}{1.2}
\begin{center}
\begin{tabular}{ |m{3.46cm}|m{1.3cm}|m{1.3cm}|m{1.3cm}|m{1.3cm}|m{1.3cm}|m{1.3cm}|m{1.3cm}|m{1.3cm}|  }
\hline
\multirow{2}[1]{*}{\centering Methods }  & \multicolumn{2}{c|}{AA} & \multicolumn{2}{c|}{LA-blood} & \multicolumn{2}{c|}{LV-blood} & \multicolumn{2}{c|}{LV-myo} \\
\cline{2-9}
&~~~ Dice & ~~~ ASD & ~~~ Dice & ~~~ ASD & ~~~ Dice & ~~~ ASD & ~~~ Dice  & ~~~ ASD \\
\hline

Seg-MRI~(ours)   & 78.3$\pm$5.5  & 15.3$\pm$10.6  & 78.9$\pm$7.1 & 15.7$\pm$6.4   & 88.5$\pm$5.2 & ~~2.9$\pm$0.5   & 75.3$\pm$3.1 & ~~2.3$\pm$2.1 \\

\hline
\hline


Seg-MRI-noDA ~(ours) & 34.8$\pm$13.8  & 20.7$\pm$13.6 & 10.6$\pm$13.8 &  19.0$\pm$8.1  & ~~5.8$\pm$4.6  & 11.2 $\pm$3.6 & 16.4$\pm$11.7 & 6.4$\pm$3.9\\

\hline
\hline
	   AdaOutput-MRI-DA  ~\cite{Tsai2018LearningTA}             & ~~~~~60.8 & ~~~~~\textbf{5.7} & ~~~~~39.8& ~~~~~8.0 & ~~~~~71.5& ~~~~~4.6& ~~~~~35.5  & ~~~~~4.6\\
	
		PnP-AdaNet-v2-MRI-DA  ~\cite{Dou2019PnPAdaNetPA}             & 43.7$\pm$10.8  & 11.4$\pm$3.2  & 47.0$\pm$7.3 & 14.5$\pm$4.1  & 77.7$\pm$10.4 & 4.5$\pm$1.4 & \textbf{48.6$\pm$2.9}  & 5.3$\pm$1.8\\
		
			SIFA-v1-MRI-DA  ~\cite{Chen2019SynergisticIA}             & ~~~~~\textbf{67.0} & ~~~~~6.2  & ~~~~~60.7& ~~~~~9.8  & ~~~~~75.1& ~~~~~4.4 &~~~~~45.8& ~~~~~4.4\\
			
			SIFA-v2-MRI-DA  ~\cite{Chen2020UnsupervisedBC}             &~~~~~65.3 &~~~~~7.3  &~~~~~62.3 &  ~~~~~\textbf{7.4} & ~~~~~78.9  & ~~~~~3.8  & ~~~~~47.3  & ~~~~~4.4 \\
				
\hline
Seg-MRI-DA~(ours)  & 66.4$\pm$28.3 & 16.9$\pm$10.8 & \textbf{63.0$\pm$34.2} & 17.1$\pm$12.9 & \textbf{79.2$\pm$12.6} & \textbf{3.5$\pm$4.4} & 47.8$\pm$8.4 &\textbf{4.1$\pm$7.3}\\

\hline
\end{tabular}
\end{center}
\label{tab:revertexpresults}
\end{table*}

\subsection{Ablation Studies}
\subsubsection{Influence of Key Component}
To evaluate the contributions of each component of our model, we perform ablation studies in this section. As described in Table~\ref{tab:ablation}, when we remove either the CDD module~($\lambda_1=0$) or CCPA module~($\lambda_2=0$) from our method, the most striking observation to emerge from the data comparison is the decrease in performance on all metrics. This indicates that both two modules play an important role in addressing the issue of domain shift in UDA.  Furthermore, when we remove these two modules~($\lambda_1=0, \lambda_2=0$), significant performance degradation occurs. This shows that enabling domain adaptation helps to improve the model generalization capability on cross-modality segmentations.

\subsubsection{Influence of Key Hyper-parameters}
We also perform a series of ablation studies to analyze the impact of different hyper-parameters on the domain adaptation performance. All models are trained from scratch with the default standard settings. Here, we study three key hyper-parameters, the trade-off ratio $\lambda_1$ and $\lambda_2$ of  $\mathcal{L}_{cdd}$ and $\mathcal{L}_{ccpa}$ in Eq.~\ref{eq:totalloss}, respectively, and the trade-off ratio $\eta$ of the Dice loss in Eq.~\ref{eq:segloss}. The results are shown in Table~\ref{tab:ablation}, consisting of three groups of experimental comparisons at the bottom of Table~\ref{tab:ablation}.

Fixing $\lambda_2 = 1, \eta=1$, we vary $\lambda_1$ ranging in [0.3, 0.6, 0.9, 1]. We first observe that as the value of $\lambda_1$ increases, the domain adaptation performance increases gradually. Compared to $\lambda_1 = 0.3$, $\lambda_1 = 1$ obtains performance improvement on all metrics, especially on the Dice of LV-blood, and the ASD of LV-myo. These tissues are difficult and contain unclear boundaries. We believe this phenomenon is because the introduced CDD module can deal well with cross-modality domain adaptation. In practice, we should set a task-specific value of $\lambda_1$ ensuring optimal results.

Similarly, we evaluate different settings $\lambda_2 =$ [0.3, 0.6, 0.9, 1] of the trade-off ratio of $\mathcal{L}_{ccpa}$ in the Eq.~\ref{eq:totalloss}. As an interesting observation, the change in the metrics is monotonic as we adjust the trade-off ratio. Additionally, $\lambda_2 = 1$ likewise achieves the best performance on all metrics, in particular, a significant improvement in the Dice of LV-blood and the ASD of LV-myo. This indicates the importance of the CCPA module in cross-modality domain adaptation segmentation.

Finally, we perform a group of experiments on the hyper-parameters $\eta$  to investigate how it affects the model performance. We report the results at the bottom of Table~\ref{tab:ablation} by adjusting $\eta$ ranging from [0, 0.2, 0.4, 0.6, 0.8, 1]. We observe in Table~\ref{tab:ablation} that our method achieves the best performance under the setting $\eta=1$ compared with other settings. Note that the cardiac structure datasets exhibit class imbalance. In this regard, it is beneficial for challenging heart segmentation tasks to combine Dice loss with cross-entropy loss. In practice, we are supposed to adjust the optimal ratio $\eta$ according to the degree of sample class imbalance in the dataset.

\subsubsection{Reverting Domain Adaptation Direction}
To investigate whether the reverse adaptation direction from CT to MRI can also be achieved and what the impact of $\mathcal{D}_s$ in domain adaptation is, we applied the same model setup, replacing the source domain as CT and the target domain as MRI for the experiments. The quantitative performance of reverting the domain adaptation direction is shown in Table~\ref{tab:revertexpresults}. Unsurprisingly, using the CT segmenter directly on MRI data also fails. Our proposed method can recover the average segmentation performances via unsupervised domain adaptation, indicating that our model has good robustness and generalization over different types of datasets and that cross-modality domain adaptation can be achieved in both directions. Interestingly, the best recovered structure in this reverse setup is the LV-blood, which is the same structure as the MRI-to-CT direction. It is also necessary to mention that the Dice of LV-blood increases from a complete failure (5.8) to a considerably higher value (79.2), and the ASD of LV-blood also shows the same trend. Compared to the adaptation from MRI to CT, the reverse direction generally yields lower performance generally, show that the difficulty of  $\mathcal{D}_s$ can affect the performance of domain adaptation, and the difficulty is not symmetric. Segmentation of cardiac MRI itself is more difficult than segmentation of cardiac CT. This is also evident from Table~\ref{tab:expresults}, where CT segmentation Dice is higher than the result of MRI in all four structures. In these respects, transferring CT segmenter to MRI seems to be more challenging.

\section{Conclusions}
In this paper, we studied the problem of unsupervised domain adaptation on cross-modality biomedical images. We introduced the CDD and CCPA modules into the model to align the latent feature space and category-centric prototype of the target domain to that of the source domain. Extensive experiments show the superior performance of our approach.




\bibliographystyle{IEEEtran}
\bibliography{ref}


\end{document}